\documentclass{article}

\usepackage{microtype}
\usepackage{graphicx}
\usepackage{subfigure}
\usepackage{booktabs} 

\usepackage{hyperref}

\usepackage[accepted]{icml2024}

\usepackage{amsmath}
\usepackage{amssymb}
\usepackage{mathtools}
\usepackage{amsthm}

\usepackage[capitalize,noabbrev]{cleveref}

\theoremstyle{plain}

\theoremstyle{definition}

\theoremstyle{remark}

\usepackage[textsize=tiny]{todonotes}

\icmltitlerunning{Transformers with Stochastic Competition  for  Tabular Data Modelling }

\begin{document}

\twocolumn[
\icmltitle{Transformers with Stochastic Competition  \\ for  Tabular Data Modelling}




\begin{icmlauthorlist}
\icmlauthor{Andreas Voskou}{cut}
\icmlauthor{Charalambos Christoforou}{cut}
\icmlauthor{Sotirios Chatzis}{cut}

\end{icmlauthorlist}

\icmlaffiliation{cut}{Cyprus University of Technology, Limassol, Cyprus}

\icmlcorrespondingauthor{Andreas Voskou}{ai.voskou@edu.cut.ac.cy}

\icmlkeywords{Tabular Data, Stochastic, Transformers, Competition}

\vskip 0.3in
]



\printAffiliationsAndNotice{}  

\begin{abstract}
Despite the prevalence and significance of tabular data across numerous industries and fields, it has been relatively underexplored in the realm of deep learning. Even today, neural networks are often overshadowed by techniques such as gradient boosted decision trees (GBDT). However, recent models are beginning to close this gap, outperforming GBDT in various setups and garnering increased attention in the field. Inspired by this development, we introduce a novel stochastic deep learning model specifically designed for tabular data. The foundation of this model is a Transformer-based architecture, carefully adapted to cater to the unique properties of tabular data through strategic architectural modifications and leveraging two forms of stochastic competition. First, we employ stochastic "Local Winner Takes All" units to promote generalization capacity through stochasticity and sparsity. Second, we introduce a novel embedding layer that selects among alternative linear embedding layers through a mechanism of stochastic competition. The effectiveness of the model is validated on a variety of widely-used, publicly available datasets. We demonstrate that, through the incorporation of these elements, our model yields high performance and marks a significant advancement in the application of deep learning to tabular data.
\end{abstract}

\section{Introduction}

 Tabular data is a fundamental and arguably one of the most commonly used formats in the fields of data science and machine learning. It is structured with rows and columns that represent individual observations and their corresponding features; this creates a simple two-dimensional, table-like body. Within it, various data types can be included. This format enjoys widespread popularity in sectors like healthcare, finance, and sciences because of its  organizational clarity and its close ties with relational databases and spreadsheets. Yet, despite its prevalence and seeming simplicity, effectively modeling tabular data for common tasks like regression or classification continues to pose significant challenges.

Features in tabular data can take several  forms, ranging from simple scalar values to custom data structures. However, in modeling scenarios, these features predominantly manifest as either continuous real values or discrete categorical variables, often encoded as positive integers. Formally, a tabular row of length $s$ can be represented as $ \boldsymbol{x} \in \mathbb{R}^{s_r} \times \mathbb{N}^{s_n}$, where $s= s_r + s_n$. Here, $s_r$ defines the number of continuous features, while $s_n$ enumerates the categorical ones.  Additionally,  the positioning of features in a tabular row holds no intrinsic geometrical meaning. Thus, we presume no inherent relations between features, in contrast to other popular data forms like images or language.

Tree-based models and particularly GBDTs \cite{friedman2002stochastic}, have long been favored for these tasks. Yet, deep learning, which has revolutionized other data realms, hasn't become the first-line approach for tabular data, a trend that is currently changing. Recent advances have seen deep learning models, especially Transformers \cite{vaswani2017attention}, outshine GBDTs in various tabular datasets. These models have proven to be very effective in tackling tabular data as they can dynamically adjust feature influences via attention mechanisms, with several successful applications \cite{gorishniy2021revisiting,arik2021tabnet} hinting at a shift towards deep learning as a leading approach in tabular data analysis.

The motivation for further research into the application of deep learning on tabular data stems also from properties beyond the potential for improved raw predictive power. In contrast to gradient boosted decision trees and similar methodologies, which have predictive capabilities that are largely fixed post-training and allow  only  for limited tuning on new data, deep networks provide inherent flexibility for continuous adaptation. This is key for techniques such as transfer and meta-learning, where knowledge from one domain can be transferred to another, significantly improving training time and overall results for new tasks. Additionally, an NN, or just its first layers, can act as a sophisticated feature extractor, making it applicable to complex tasks beyond the initial training or for further secondary analysis.

In this paper, we delve further into the field and propose a novel Deep Learning architecture tailored for Tabular Data.  Particularly, drawing on recent advancements, we adopt the Transformer encoder \cite{vaswani2017attention} as our foundational architecture. We then introduce substantial modifications to adapt it to tabular data, introducing a hybrid transformer encoder layer. This layer, in addition to the standard components, includes (i) a parallel fully connected feature aggregating element and (ii) an attention bias term. In addition, we infuse sophisticated stochastic competition techniques into the model. These include the utilization of the powerful stochastic "Local Winner Takes All" (LWTA) layer \cite{panousis2019nonparametric}, which has shown exceptional results in various contexts, though it has not yet been applied to tabular data; and a novel Mixture Embedding layer for numerical input features, enabling a more comprehensive feature representation.

The remainder of this paper is organized as follows. The next Section offers an overview of related work. In Section 3, we introduce the proposed approach, explain its main architectural assumptions and components, and derive the training and inference algorithms. Section 4 provides a deep experimental evaluation of our proposed approach, using established benchmarks in the field; this is combined with a long ablation study. Finally, in Section 5, we conclude this paper by drawing some key insights.

\section{Related Work}

As previously outlined, the most established methods in Tabular Data Modeling (TDM) currently belong to the family gradient-boosted decision trees (GBDT). These algorithms rely on an ensemble of weak learners, sequentially generated as corrections to the existing ensembles in a gradient-driven fashion. The most renowned and popular variants of such algorithms include Catboost\cite{prokhorenkova2018catboost}, XGBoost\cite{chen2015xgboost}, NGboost\cite{duan2020ngboost}, and LightBoost\cite{ke2017lightgbm}. The popularity of these approaches, stems from their high performance and ease of use.


Until the close of the previous decade, methodologies for deep learning with Tabular Data primarily focused on multi-layer perceptrons (MLPs) and similar basic architectures. While  MLPs remain effective in certain contexts, provided they receive appropriate treatment\cite{gorishniy2021revisiting,NEURIPS2021_c902b497}, recent years have witnessed a surge in more sophisticated neural network designs, yielding remarkable results. These contemporary designs have adopted diverse strategies, including emulating decision trees or other types of weak learners drawing inspiration from GBDTs. Two seminal architectures embodying this philosophy are NODE\cite{popov2019neural} and GrowNet\cite{badirli2020gradient}.

While these methodologies have recorded commendable outcomes, the trajectory in recent research has been the inclination towards Transformer-based architectures.
Examples like TabNet\cite{arik2021tabnet} harness the power of the Transformer and the attention mechanism, giving strong results through an encoder-decoder framework. Conversely, TabTransformer\cite{huang2020tabtransformer} deploys the transformer to process categorical tabular features and subsequently amalgamates the resultant representations with a fully connected layer to address the numerical features. FtTransformer\cite{gorishniy2021revisiting}, meanwhile, employs an encoder-only design to analyze all features and projects individual categorical and numerical features into distinct vector representations using a simple yet effective linear embedding layer. Furthermore, SAINT \cite{somepalli2021saint} goes beyond row-by-row processing through the addition of an inter-row attention layer. The literature extends further with many notable works such as \citet{zhu2021converting} and \citet{hollmann2022tabpfn}.

Apart from proposing sophisticated network architectures, a number of studies have investigated the implications tied to distinct attributes and settings that underpin deep learning practices. Characteristic studies in this context consider pretraining  \cite{rubachev2022revisiting,iida2021tabbie}, as well as various embedding approaches that yield strong results even with simple architectures such as MLP-PLR \cite{gorishniy2022embeddings}.  Standout contributions include \citet{kotelnikov2023tabddpm}, which employs stable diffusion probabilistic models, and \citet{gorishniy2023tabr}, which employs retrieval augmentation strategies.

\section{The model}

 \subsection{Overview}

\begin{figure}[h]
\begin{center}
\includegraphics[scale=0.60]{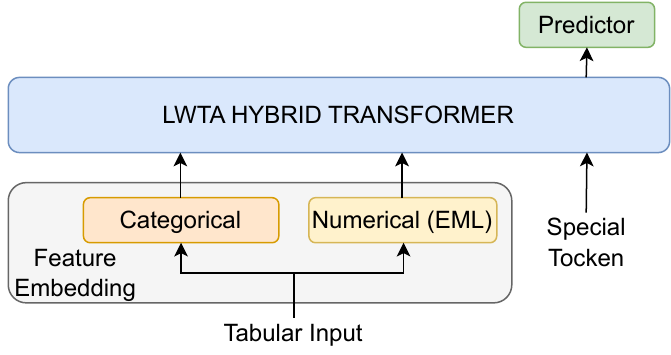}
\end{center}
\caption{Overview of our approach, exhibiting its core modules.}
\label{fig:overview}
\end{figure}

In Figure \ref{fig:overview}, we provide a comprehensive overview of the proposed model, which employs a hybrid architecture grounded on an encoder-only Transformer. This foundational architecture is augmented with stochastic elements and additional structural modifications, which we  discuss in greater detail later in this Section. 

Our proposed adaptations do not obliterate the necessity for a specific input structure compatible with the standard Transformer encoder.
To achieve compatibility  our first step is to adapt the original data format, defined in $\mathbb{R}^{s_r} \times \mathbb{N}^{s_n}$, to one that fits the Transformer. Through  embedding layers, each feature $x_i,   i\in \{1,..,s\} $, be it numerical or categorical, is mapped onto a d-dimensional representation vector, given by $\boldsymbol{h}_i \in \mathbb{R}^d$. Eventually, a given input $\boldsymbol{x}=(\boldsymbol{x}_i)_{i=1}^{s}$ is mapped to a vector sequence $ \boldsymbol{h} \in \mathbb{R}^{d\cdot s}$.
Alongside this representation, we also add a vector, $\boldsymbol{h}_{special} \in \mathbb{R}^d$, that corresponds to an artificial "special token" with  static  value. The terminal representation of this token is fed to a final regression or classification head, depending on the  task.

While our architectural design shares similarities with usual Transformers and  preceding models on tabular data and other domains, it distinguishes itself through three key innovations that enhance its predictive capability:
i) The adoption of the sophisticated stochastic LWTA layer \cite{panousis2019nonparametric}.  The latter has been shown to yield improved results in a wide range of applications; yet, it has never been employed to networks designed for Tabular Data.
ii) The introduction of a novel data-driven probabilistic  selection among alternative (linear) feature embeddings. This enhancement adds an extra element of stochasticity and promotes richer feature representations.
iii) The introduction of the Hybrid Transformer module, which is specifically designed for tabular data. This module merges the core Transformer encoder layer architecture with a parallel fully connected aggregation module. Tailored to capitalize on the static structure of tabular data, this aggregation module works by projecting the hidden representations back to scalar values and processing the aggregate result.

In the following subsections, we elaborate on each of the core novel elements that compose our proposed approach.

\subsection{Local Winner Takes All}

The Stochastic 'Local Winner Takes All' (LWTA) layer, as introduced by Panousis et al. (2019), is a sophisticated alternative to common deterministic layers. It enhances performance by incorporating stochasticity and yielding rich, sparse representations. An LWTA layer consists of linear units and introduces nonlinear behavior through stochastic competition within blocks of layer neurons, rather than relying on deterministic activation functions such as ReLU. Within each block, only one neuron, termed the 'winner,' is activated based on probabilistic selection. All other neurons remain inactive, transmitting zero values.

In a more formal notation, let us consider  the input and output vectors of a typical linear layer, denoted by $\boldsymbol{x} \in \mathbb{R}^J$ and $\boldsymbol{y} \in \mathbb{R}^H$ respectively, with the associated weight matrix denoted as $W \in \mathbb{R}^{J\times H}$. In the LWTA approach, the elements of $\boldsymbol{y}$ are partitioned into $K$ distinct, non-overlapping blocks, each containing $U$ elements. Concurrently, the weight matrix $W$ is restructured into $K$ separate submatrices. This gives us $\boldsymbol{y}_k \in \mathbb{R}^U$ and $W_k \in \mathbb{R}^{J \times U}$ for each block $k \in \{1,2,\dots,K\}$. Within each block, the output values compete against one another and  only one, the "winner", is retained; the remaining elements are set to 0. The aforementioned competition is technically implemented as a stochastic sampling process inside each block. In this process, the winner indicator, a latent one-hot vector, $\boldsymbol{\xi}_k \in \mathbf{onehot}(U)$ is sampled from a  discrete  posterior $D(\boldsymbol{\xi})$. The posterior  logits are  directly proportional to the linear computations of each respective unit, passed through a softmax. The final layer   output   $\boldsymbol{y}_k$ for the block $k$ is gained by using the postulated  $\boldsymbol{\xi}_k$ in a simple masking operation as in (\ref{eq:lwta_simplified}).
\begin{equation}
\label{eq:lwta_simplified}
\boldsymbol{y}_{k} =     \boldsymbol{\xi}_{k}\odot ( {W}_k \boldsymbol{x} ), \; {\xi}_k \sim  \mathrm{D}\bigg(\boldsymbol{\xi}_k \bigg|  W_k \boldsymbol{x}    \bigg),\;  \; \forall  k \in \{1,2,..K\}
\end{equation}
where $\odot$ stands for element-wise multiplication. During training, $\boldsymbol\xi_k $ is approximated via a Gumbel-Softmax differentiable sample \cite{jang2016categorical}, to ensure  effectiveness and stability:
\begin{equation}
\begin{split}
\boldsymbol{\xi}_k &= \dfrac{\exp{{((  \boldsymbol{\eta}_k +  \boldsymbol{g}_k})/{T})}}{\sum_{i=1}^U{exp{(({{\eta_{k,i}} + g_{k,i})}/{T})}} } \\ 
\boldsymbol{g} &= - \log(-\log{\boldsymbol{z}}), \;  \boldsymbol{z} \sim \mathrm{U}(\boldsymbol{0},\boldsymbol{1}) 
\label{eqn:elbo_full}
\end{split}
\end{equation}
where $\boldsymbol{\eta}_k= W_k \boldsymbol{x}$, and $T$ is a  temperature hyperparameter.

The motivation for employing this approach stems primarily from its considerable success across various tasks and frameworks, such as image classification \cite{panousis2022competing}, meta-learning \cite{kalais2022stochastic}, and its use in transformer-based models for sign language translation across different languages \cite{Voskou_2021_ICCV,gueuwou2023afrisign,voskou2023new}, among others. However, this approach has not yet been applied to tasks involving tabular data. Moreover, the LWTA has emerged as a particularly effective method for preventing overfitting without the addition of extra parameters.

\subsection{Feature Embedding - Embedding Mixture Layer } 

Feature embedding serves as a pivotal element in models like the one we propose, acting as the bedrock upon which later processing stages are built. In our approach, each categorical feature is separately processed via a standard linear embedding layer. This technique is  stable and well-grounded in the literature, sharing conceptual similarities with word embedding commonly used in NLP.


\begin{figure}
    \centering
    \includegraphics[scale=0.71]{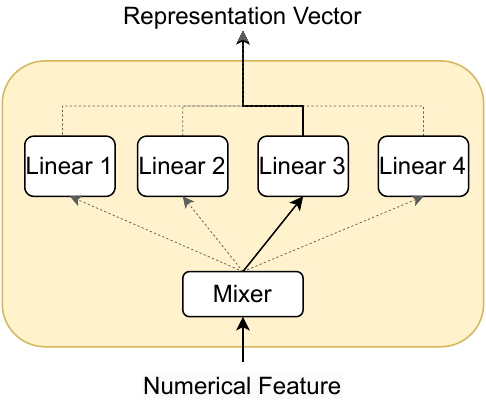} 
    \caption{The embedding mixture layer}
    \label{fig:NewLayersEM}
\end{figure}
Embedding of continuous  values is much underexplored. Earlier work \cite{gorishniy2021revisiting}  has mostly been limited to simple linear projections, computed independently for each feature. Recently,  non-linear approaches have been explored and proved to be beneficial to the predictive accuracy \cite{gorishniy2022embeddings}.
In this work, we progress one step further, proposing a  novel stochastic embedding layer that improves the expressive power of the vanilla approach.  In our proposed method, instead of having a single pair of weight and bias vectors, we use a set of  $J$ such pairs, defining $J$ alternative (linear) embeddings, each indicated by an indicator $j$.  To gain the representation vector of a continuous input, $x_i$, the model has now to select one of the so-defined alternative linear projections. It does so in a stochastic manner, where the probability of one alternative embedding being selected  is  driven by the value of $x_i$ via (\ref{eq:embed}); this selection rationale is illustrated in Figure \ref{fig:NewLayersEM}. 
\begin{equation}
\label{eq:embed}
f_{emb}(x_i ) =  x_i  \boldsymbol{w}_j + \boldsymbol{b}_j, \;\; \;\;  \; j \sim \boldsymbol{P}(\cdot|x_i,\boldsymbol\theta_w,\boldsymbol\theta_b) 
\end{equation}
where the posterior probability distribution over the linear mapping  reads
\begin{equation}
\boldsymbol{P}(j|x,\boldsymbol\theta_w,\boldsymbol\theta_b)   = \frac{e^{t_j}}{\sum_{j=1}^{J} e^{t_j}}, \;\; \;\;  \boldsymbol{t}= x_i \cdot \boldsymbol{\theta}_w + \boldsymbol{\theta}_b
\end{equation}  
with $\boldsymbol{\theta}_w ,  \boldsymbol{\theta}_b  \in \mathbb{R}^{J}$ denoting the trainable parameters directly involved in the selection process. 

This embedding selection scheme can be described as a sort of competition among sub-parts at the embedding layer level; each competitor aims to dominate a broader range of input values. We posit that,  in this way, the embedding engine can produce representations that are significantly richer than a single linear mapping. The eventually obtained embedding vector can vary considerably more than vanilla embeddings, based on the value regions of the input feature; this may allow for the identification of behavioral changes and shifts in statistical importance related to that feature. Additionally, the induced probabilistic transitions between different linear embedding enhance accuracy in uncertain areas of mapping, and also  reduce the risk of over-fitting.


\subsection{Hybrid Transformer module}

\begin{figure}
    \centering
    \includegraphics[scale=0.69]{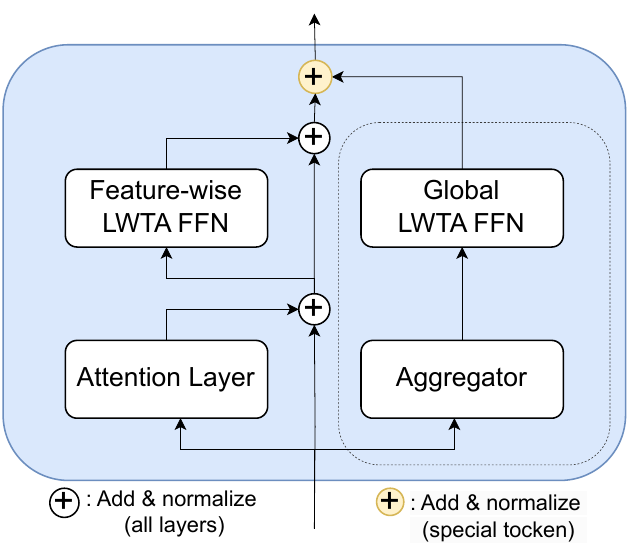}
    \caption{Illustration of a single Hybrid Transformer layer.}
    \label{fig:NewLayersHy}
\end{figure}


Typical Transformer input modalities, like text and videos, frequently display dimensionality that is subject to change, such as sentence lengths or video duration. Conversely, tabular datasets exhibit fixed, predefined dimensions. This distinct property offers an avenue for integrating static elements into the network, which would be unattainable in dynamically changing contexts.

Our so-obtained hybrid Transformer layer melds two essential subcomponents.  Similar to a standard Transformer encoder layer, the first component is a feature-wise sequential arrangement of a Self-Attention layer and a Fully-Connected layer. In our work, we enhance the attention dot-product with a bias term, an adjustment we have empirically found to be subtle yet effective. While the incorporation of various types of bias in attention has been previously explored, such as in \citet{duffer2022} where it was used to add relative positional information, our application is the first aimed at leveraging the structural properties of tabular data. The second component, a novel aspect of our design, is a \emph{parallel module}. This module can technically be described as an LWTA-based global feedforward layer, as illustrated in Figure \ref{fig:NewLayersHy}. This innovation is inspired by previous research \cite{gorishniy2021revisiting,NEURIPS2021_c902b497,shwartz2022tabular} showing that despite the popularity of Transformer architectures, fully connected architectures can still yield remarkable results and should not be overlooked. Our hybrid approach facilitates an effective blend of static and dynamic feature interactions, contrasting with the purely dynamic nature of typical Transformers. Through this modification, we enhance the model's predictive capability with only a small increase in computational cost.

The new module is presented with the $d$-dimensional embedding of each of the $s$ input features, reprojects them onto scalar values and aggregates them into a single $s$-dimensional representation vector through the operation $\Phi : \mathbb{R}^{d\cdot s} \rightarrow \mathbb{R}^s$:
\begin{equation}
\Phi(\mathbf{h}) = \left( \boldsymbol {w}_i \cdot \boldsymbol{h}_i + b_i \right)_{i=1}^{s} \; \text{where} \;      \; \boldsymbol{w}_i,\boldsymbol{h}_i \in \mathbb{R}^d,\; b_i \in \mathbb{R}
\end{equation}
The so obtained  vector, $\Phi(\mathbf{h})$, is presented to a subsequent LWTA layer, followed by a Linear layer; this yields an output vector $\boldsymbol{z} \in \mathbb{R}^d $.  The output from this module is incorporated into the  representation of the special token, in an additive (residual) manner.


\subsection{Training and Inference}


The training objective of  our proposed model  is  formulated as follows:
\begin{align}
\begin{split}
\mathcal{L}(\phi) =  & \mathbb{E}_{q(\cdot)}\big[\log p(\mathcal{D}|\{\phi\}) \big]  - \mathrm{KL} \big[Q(\{ \boldsymbol \xi \})||P(\{ \boldsymbol \xi \})\big]\\& -\mathrm{KL} \big[Q(\{ j\})||P(\{j\})\big]
\end{split}
\label{eqn:elbo}
\end{align}
where $ \{ \boldsymbol \xi \}$ the set of the LWTA winner indicators,  $ \{ \boldsymbol j \}$ the embedding selection indicators  and $\{\phi\}$ represents all the trainable parameters. It is captured by a composite functional consisting of three terms. The first term corresponds to the primary  objective. It incorporates the standard crossentropy loss for classification tasks and the mean squared loss for regression scenario. In both cases, the latent indicator vectors $\boldsymbol{\xi}$ and $\boldsymbol{j}$ are replaced by a differentiable (reparameterized) expression obtained through the Gumbel-Softmax trick. The second term encapsulates the Kullback-Leibler divergences between the posteriors and the priors of the  winner indicators, using a uniform discrete prior distribution $U$:
\begin{equation}
\mathrm{KL}[Q(\boldsymbol{\xi})||P(\boldsymbol{\xi})] =\sum\limits_{\forall \boldsymbol{\xi}}{\sum\limits_{i=1}^U{Q(\xi_i)\log{(Q(\xi_i)/U_i)} }}  
\end{equation}
The third term is similar to the second, but quantifies the KL divergence between the posterior of embedding selection and a uniform discrete prior.

For model evaluation and inference, predictions are gained via Bayesian averaging. By executing the model multiple times, we average the resultant outputs from the employed classification or regression head.

\section{Experimental Results}

\subsection{Benchmarking datasets}
\begin{table*}[t]
\caption{Key statistics and properties of benchmarking datasets.}
\label{tab:datasets}
\begin{center}
\begin{tabular}{lllllllll}
\toprule
\multicolumn{1}{c}{} & \multicolumn{1}{c}{\bf HI} & \multicolumn{1}{c}{\bf AD} & \multicolumn{1}{c}{\bf OT} & \multicolumn{1}{c}{\bf HE} & \multicolumn{1}{c}{\bf JA} & \multicolumn{1}{c}{\bf YE} & \multicolumn{1}{c}{\bf DI} & \multicolumn{1}{c}{\bf HO} \\
\midrule
Total Entries & 98049 & 48842 & 61878 & 65196 & 83733 & 515345 & 53940 & 22784 \\
Total Features & 28 & 14 & 93 & 27 & 54 & 90 & 9 & 16 \\
Catg Features & 0 & 8 & 0 & 0 & 0 & 0 & 3 & 0 \\
Task (Classes)  & C(2) & C(2) & C(9) & C(100) & C(4) & R & R & R \\
\bottomrule
\end{tabular}
\end{center}
\end{table*}
In the experimental section, we employ 8 publicly available tabular datasets, in the same form as previously utilized in analogous research, such as \citet{gorishniy2021revisiting}, and \citet{gorishniy2023tabr}. We use exactly the same train-validation-test split to facilitate fair comparison. Specifically, our analysis involves two datasets for binary classification, namely Higgs Small(HI) and Adult(AD); three datasets designed for multi-class classification, namely Otto Group Products(OT) with nine classes, Helena(HE) with 100 classes, and Jannis(JA) with four classes; and three datasets tailored to regression tasks, namely Year Prediction(YE), Dimanond(DI), and House16H(HO). As reference metrics, we follow a common practice and  use MSE for Regression and Accuracy for Classification Tasks.
The bulk of the selected datasets are medium-sized, with row counts ranging from 20,000 to 100,000. However, to also examine how performance changes when using a significantly larger dataset, we also use Year Prediction, a particularly popular dataset encompassing around half a million features. In the context of feature types, the majority of datasets include numerical attributes, with feature dimensions ranging from 5 to 93. Exceptions to this pattern are  Adult and Diamond, which additionally incorporate categorical features. Detailed analysis of data statistics is provided in Table \ref{tab:datasets}. 

\subsection{Experimental setup}

In all experiments, the AdamW \cite{loshchilov2017decoupled} optimization algorithm was selected, with a small weight decay rate, \( wd \leq 10^{-4} \). Training was divided into two sequential phases: a short warm-up  featuring an ascending learning rate, and a subsequent main training  part. In the latter phase, the learning rate commenced at \( lr = 10^{-3} \) and was subject to a 50\% reduction upon reaching a performance plateau. For LWTA layers, a fixed block size of \( U = 2 \) is employed, while for the embedding mixture layers, \( J = 16 \) is used, both supported by preliminary analyses.
 Additional hyper-parameters are an 8 head multi-head attention; an mc-dropout rate of \( p = 0.1-0.25 \); and a Gumbel Softmax temperature \( T = 0.69 \) for training and \( T = 0.01 \) for inference. As usual with Gumbel-Softmax reparameterization, it suffices that we consider sample size $N=1$ for training; we draw $N=64$ samples for inference. For input data prepossessing, appropriate normalization/scaling was employed, except for the OT dataset where original scaling was retained as suggested by \citet{gorishniy2021revisiting}. Additionally, we re-scale the labels of HO and DI by a factor of $10^{-4}$ and $10^2$, respectively for better illustration purposes.  All reported results regarding the proposed method correspond to the average of 4 different trainings from different random seeds; all ensemble scores are combinations of these 4 runs. 

\subsection{Results Discussion}

\begin{table*}
\caption{Results comparison with related Deep Neural Networks.}
\label{tab:single_models}
\begin{center}
\begin{tabular}{llllllllll}
\toprule
\multicolumn{1}{c}{\bf } & \multicolumn{5}{c}{\bf Classification($ \text{Acc}\uparrow$)}  & \multicolumn{3}{c}{\bf Regression($ \text{MSE}\downarrow$)} &  \multicolumn{1}{c}{ avg.} \\
\multicolumn{1}{c}{\bf Model} & \multicolumn{1}{c}{\bf HI} & \multicolumn{1}{c}{\bf AD} & \multicolumn{1}{c}{\bf OT} & \multicolumn{1}{c}{\bf HE} & \multicolumn{1}{c}{\bf JA} & \multicolumn{1}{c}{\bf YE} & \multicolumn{1}{c}{\bf DI} & \multicolumn{1}{c}{\bf HO}  &    \multicolumn{1}{c}{ rank}  \\
\midrule
MLP & 71.9\% & 85.3\% & 81.6\% &  38.3\% & 71.9\% & 78.37 & 1.960 & 9.6845 & 4.1 \\
MLP-PLR & 72.9\% & \textbf{87.0\%} & 81.9\% & -- & -- & -- &\textbf{1.800}&   \textbf{9.339} & 1.7 \\
Node & 72.6\% & 85.8\% & -- & 35.9\% & 72.7\% & 76.40 & -- & --  &  3.8 \\
FtTransformer & 73.0\% & 85.9\%& 81.7\% & 39.1\% & 73.2\% & 78.40 & -- & 10.480  & 3.0 \\ 
\midrule
STab & \textbf{73.2\%} & 86.1\%& \textbf{82.5\%} & \textbf{39.4\%} & \textbf{73.6\%} & \textbf{76.10} & 1.825 & 9.650 & 1.4 \\
\bottomrule
\end{tabular}
\end{center}
\end{table*}

Table \ref{tab:single_models} presents a comparative evaluation of our proposed model against leading deep-learning benchmarks, specifically MLP-PLR, NODE, FtTransformer, and SAINT, as well as a basic MLP. To maintain a focused examination of architectural differences, we intentionally exclude methods that rely on transfer learning or data augmentation. For the proposed model (STab)\footnote{https://github.com/avoskou/Transformers-with-Stochastic-Competition-for-Tabular-Data-Modelling}, we employ our recommended hyperparameters obtained though a brief tuning procedure and empirical consideration. For established benchmarks, we cite results from existing literature as provided in \citet{gorishniy2021revisiting,rubachev2022revisiting,gorishniy2023tabr}, or \citet{somepalli2021saint}.  It is important to mention that all the reported third-party results are the outcome of  well-conducted hyperparameter tuning, typically more extensive than ours, and the reported numbers have been verified. This approach conserves computational resources and ensures impartiality by relying on multi-party verification of performance metrics.

As shown in Table \ref{tab:single_models}, all high-performing models display similar levels of performance, with none exhibiting significant superiority. This suggests that the datasets might be nearing the optimal results achievable, making further improvements challenging; even a slight edge could be crucial, particularly in competitive settings. Our model shows superior performance, surpassing existing neural network architectures in 5 out of the 8 evaluated benchmarks. Exceptions are observed in the HO, AD, and DI datasets, where our model remains competitive, ranking second and only behind MLP-PLR. Notably, these exceptions are datasets with fewer features, with DI and AD also being the only two that include categorical features. This highlights a specific advantage of our approach in handling datasets that have a larger number of features and are exclusively composed of numerical features.  Furthermore, we observe that the deviation in scores across different random seeds is relatively low, especially given the stochastic nature of our approach, with $\sigma^2 < 0.2\%$ for all classification tasks, $\approx 0.2$ for YE, and $\approx 0.025$ for DI; the exception is the HO dataset, where $\sigma^2 > 0.5$, which is high relative to its mean value, likely reflecting its highly noisy nature.

\begin{table*}
\caption{ Ensemble models  results comparison with Deep Networks and Gradient Boosted Decision Trees }
\label{tab:ensemble_models}
\begin{center}
\begin{tabular}{lllllllll}
\toprule
\multicolumn{1}{c}{\bf } & \multicolumn{5}{c}{\bf Classification($ \text{Acc}\uparrow$)} & \multicolumn{3}{c}{\bf Regression($ \text{MSE}\downarrow$)}   \\
\multicolumn{1}{c}{\bf Model} & \multicolumn{1}{c}{\bf HI} & \multicolumn{1}{c}{\bf AD} & \multicolumn{1}{c}{\bf OT} & \multicolumn{1}{c}{\bf HE} & \multicolumn{1}{c}{\bf JA} & \multicolumn{1}{c}{\bf YE} & \multicolumn{1}{c}{\bf DI} &\multicolumn{1}{c}{\bf HO} \\
\midrule 
XGBoost & 72.6\% & \textbf{87.2\%} & 83.0\% & 37.5\% & 72.1\% & 79.98 & 1.877 & 10.09 \\
$\text{XGBoost}_{ens}$ & 72.8\%& \textbf{87.2\%} &\textbf{83.2\%} & 38.8\% & 72.4\% & 78.49 & 1.850 & 10.00 \\
CATBoost & 72.6\% & 87.1\% & 82.5\% & 38.5\% & 72.3\% & 78.98 & 1.796 & 9.720 \\
$\text{CATBoost}_{ens}$ & 72.9\%&\textbf{87.2\%} & 82.7\% & 38.5\% & 72.7\% & 78.11 & \textbf{1.769} & 9.645 \\
\midrule
$\text{MLP-PLR}_{ens}$ & 73.5\% &\textbf{87.2\%} & 82.2\% & -- & -- & -- & 1.769 & \textbf{8.958} \\
$\text{Node}_{ens}$ & 72.7\% & 86.0\% & -- & 36.1\% & 73.0\% & 76.02 & -- & -- \\
$\text{FtTransformer}_{ens}$ & 73.3\% & 86.0\% & 82.4\% & 39.8\% & 73.9\% & 76.51 & -- & 10.17 \\
\midrule
$\text{STab}_{ens}$ & \textbf{73.6\%} & 86.2\%&\textbf{83.2\%} &\textbf{40.0\%} & \textbf{74.0\%} &\textbf{75.60} & 1.781 & 9.300 \\
\bottomrule
\end{tabular}
\end{center}
\end{table*}

Beyond the main results of Table \ref{tab:single_models} in Table \ref{tab:ensemble_models}, we extend the comparison to include ensemble models as well as two established GBDT paradigms in both single and ensemble configurations. While our model's superiority persists in ensemble settings, the margin of lead narrows slightly.   Gradient Boosting models in their ensemble form   closely align with our results on   the OT task, and CATBoost' s marginally outperform us on DI. In addition, our model seems to benefit slightly less from ensembling compared to some older deterministic deep networks, possibly due to its inference mechanism via Bayesian averaging.  
Nonetheless, the ensemble version of our model remains the state-of-the-art solution for the majority of the evaluated tasks.

\subsection{Ablation study}


\begin{table*}[ht]
\caption{ Ablation study on different model variants.  }
\label{tab:variants_comp}
\begin{center}
\begin{tabular}{l lllll lll}
\toprule
\multicolumn{1}{c }{\bf Transformer } & \multicolumn{5}{c }{\bf Classification($ \text{Acc}\uparrow$)} & \multicolumn{3}{c}{\bf Regression($ \text{MSE}\downarrow$)}   \\
\multicolumn{1}{c}{\bf Variant} & \multicolumn{1}{c}{\bf HI} & \multicolumn{1}{c}{\bf AD} & \multicolumn{1}{c}{\bf OT} & \multicolumn{1}{c}{\bf HE} & \multicolumn{1}{c}{\bf JA} & \multicolumn{1}{c}{\bf YE} & \multicolumn{1}{c}{\bf DI} & \multicolumn{1}{c}{\bf HO} \\
\midrule 
Vanilla  & 73.0\% & 85.9\%& 81.7\% & 39.1\% & 73.2\% & 78.40 & 1.89 & 10.48  \\
Stochastic  & 73.1\% & 86.0\% & 82.1\% & 39.3\% & 73.4\% & 77.05   & 1.84 & 10.02 \\
Hybrid  & \textbf{73.2\%}  & 86.0\% & 81.9\% & 39.0\% & 73.2\% & 76.75 & 1.87 & 10.08 \\
Full-model & \textbf{73.2\%} & \textbf{86.1\%} & \textbf{82.5\%} & \textbf{39.4\%} & \textbf{73.6\%} & \textbf{76.10} & \textbf{1.83} & \textbf{9.65} \\
\bottomrule
\end{tabular}
\end{center}
\end{table*}
\subsubsection{On the impact of the proposed modules. }

In Table \ref{tab:variants_comp}, we present a comparative analysis between variants of our approach, aiming to examine the impact of each proposed element. The vanilla variant is a regular transformer encoder that incorporates neither the task-specific architectural modifications (the parallel module and the attention bias) nor the stochastic competition elements (embedding mixture and LWTA); this is equivalent to the FtTransformer model. The subsequent two cases correspond to the implementation of only the stochastic competition modules (Stochastic) and the use of the Hybrid Transformer Layer in a deterministic setup, similar to the vanilla version(Hybrid). Finally, we include the results of the full model to facilitate easier comparison.

Upon examination, it is evident that both the Stochastic and Hybrid variants exhibit performance enhancements over the Vanilla model; these enhancements have been obtained independently  but the cumulative effect is more prominent when combined under the full model configuration. However, notable exceptions exist, such as in the case of the JA and HE datasets, where the Hybrid architecture, when applied in isolation, either fails to offer any benefit or reduces the performance. Similarly, for the HI dataset, the full model does not manifest any marked advantage over the deterministic hybrid framework.

\subsubsection{Analysing  competition parameters $U$$\And$$J$. }
\begin{table}[ht]
\caption{ Targeted study on the the effect of mixture embedding  parameter $J$ (upper) and LWTA block size $U$ (lower).}
\label{tab:emlJ}
\begin{center}
\begin{tabular}{l  ll  ll}
\toprule
\multicolumn{1}{c}{\bf } & \multicolumn{1}{c}{\bf HI($\uparrow$)} &\multicolumn{1}{c}{\bf HE($\uparrow$)} & \multicolumn{1}{c}{\bf DI($\downarrow$)}&\multicolumn{1}{c}{\bf HO($\downarrow$)}\\
\midrule 
J= 64& 73.2\% & 39.4\% & 1.84& 9.94 \\
J= 16& 73.2\% & 39.4\% &1.83& 9.65 \\
J= 4& 73.3\% &  39.4\% & 1.84& 9.67 \\
J= 1&  73.2\% & 39.1\% & 1.87& 9.88 \\
\midrule
U= 4& 73.1\% & 39.2\% & 1.83& 9.70 \\
U= 2& 73.2\% &39.4\% & 1.83& 9.65 \\
\bottomrule
\end{tabular}
\end{center}
\end{table}

To evaluate the influence of the Probabilistic Embedding Mixture and the relevant parameter \(J\) (mixture components) on our model's performance, we conducted a specific study. The results are displayed in the upper section of Table \ref{tab:emlJ}, focusing on the significance of the parameter \(J\). Notably, setting \(J=1\) is equivalent to employing a standard linear embedding. Data from Table \ref{tab:emlJ} suggest that, in many cases, \(J=16\) is the optimal value or closely approximates it. Moreover, there is a noticeable improvement compared to the standard linear numerical feature embedding. However, minor adjustments to \(J\), whether below or above the optimal, typically do not lead to significant shifts in performance metrics. This finding supports our recommendation of a fixed \(J=16\) for the main analysis, thus diminishing the need for further time-consuming tuning.

Similarly, to further justify our decision to maintain a constant LWTA block size of \(U=2\), beyond previous literature, we conducted a brief analysis on the impact of a larger block size (\(U=4\)) in the lower section of Table \ref{tab:emlJ}. The findings indicate that increasing the block size usually results in either suboptimal performance or only a minimal impact. These results, in line with prior studies, further validate our selection of \(U=2\) as an effective default setting, reducing the incentive for additional exploration.

\subsubsection{Bayesian Averaging and Sample  Size}
\begin{figure}
    \centering
    \includegraphics[scale=0.37]{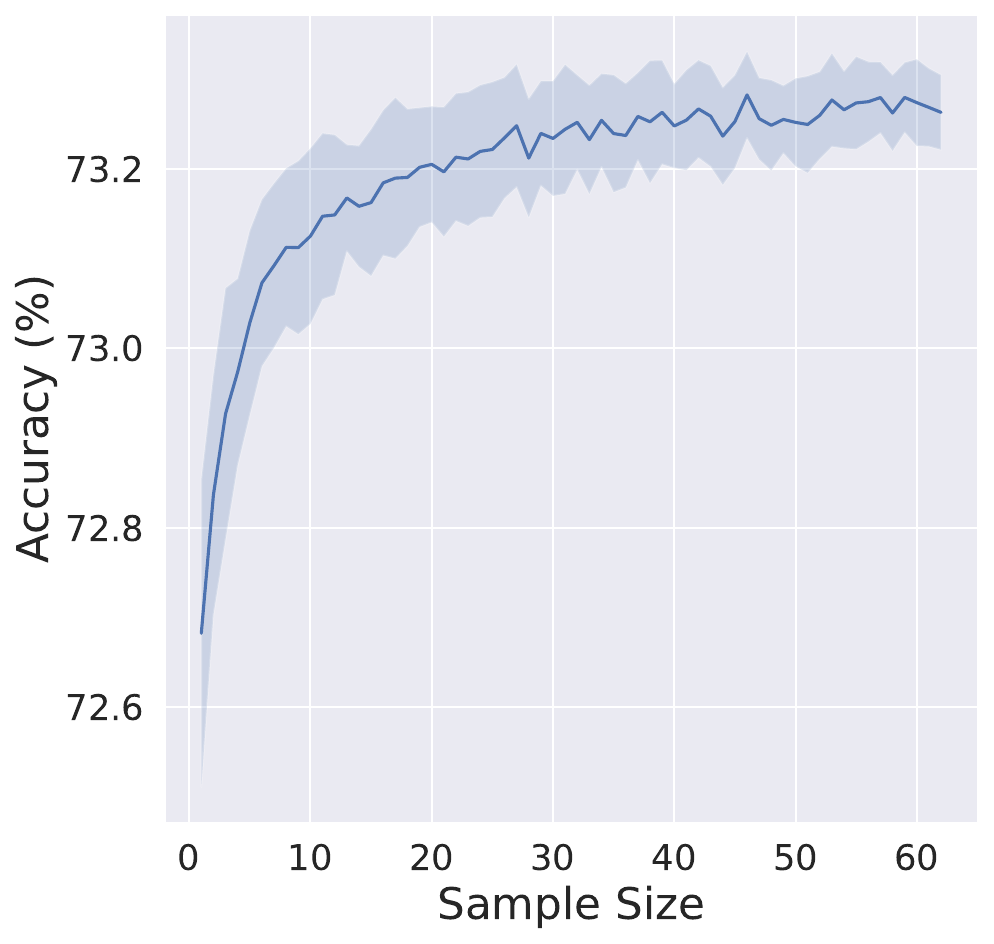}
        \includegraphics[scale=0.37]{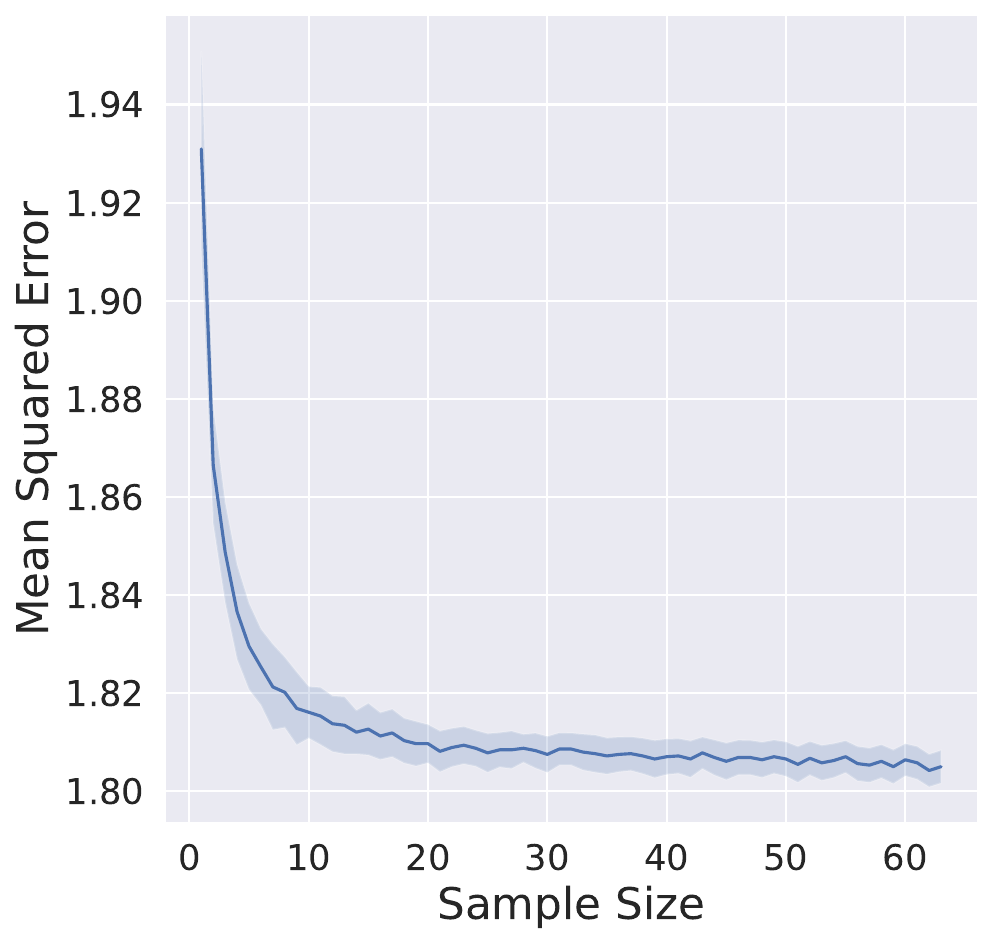}

    \caption{The effect of Sample size N on model's performance : (Top) Higgs  , (Bottom) Diamond  }
    \label{fig:BAver}
\end{figure}
Due to the inherent stochastic nature of our model, we obtain its final prediction through Bayesian averaging. In Figure \ref{fig:BAver}, we examine the relationship between sample size and prediction quality. As  expected, we find that increasing the sample size generally improves  and stabilizes the prediction, which yet starts reaching  a plateau for around  $N=20$.

While our averaging approach may  look similar to model ensembling, it's crucial to point out that they differ in key aspects. Unlike model ensembling, which requires training multiple (N) distinct models, our method needs just a single model to be trained.  This means no need for extended training processes neither additional memory and storage space. Additionally, while it is true that inference time increases linearly with N in either case, this does not hold for single-row inference or small batches. In these cases, even for large N, drawing N samples can be performed in parallel on a single GPU without additional delays. This is particularly important for real-time applications requiring low latency and rapid response times.



\section{Conclusions and future work}

In this paper, we introduce a novel approach to tabular data 
modelling by harnessing contemporary stochastic deep learning, with a particular emphasis on stochastic competition techniques. We employ a 
 Transformer-based model  with a modified task-adapted 
architecture. The model is further augmented by the 
integration of the stochastic LWTA layer. Additionally, we 
unveil a distinctive embedding mixture layer for numerical features, 
selecting among multiple linear mappings in a probabilistic fashion, through a stochastic competition mechanism. As a testament to our approach's efficacy, we 
secured state-of-the-art results on a 5 out of 8 popular 
benchmarks and achieved second place among
recent deep learning methodologies in the remaining instances. Notably, 
these advantages persist even in ensemble model configurations.

In upcoming research endeavours, we recommend a thorough exploration of stochastic competition methods, with the goal of enhancing model performance for tabular data and setting the stage for a deep learning framework in this GBDT-dominated area. Another avenue of interest is understanding how these stochastic techniques can leverage sample outcomes to estimate metrics beyond just expected values; this includes assessing uncertainties and probing into the distributional aspects of predictions. Also, incorporating advanced strategies, such as smart data augmentation, transfer learning, and meta-learning, offers a promising perspective for future studies. Historically, these methodologies have demonstrated their effectiveness by markedly improving model outcomes, suggesting their potential to elevate the efficacy of our proposed architecture.

\nocite{langley00}

\bibliography{example_paper}
\bibliographystyle{icml2024}

\newpage
\appendix
\onecolumn
\section{Appendix}

\subsection{The computational cost of the hybrid module}

\begin{table} [h]
\caption{Percentage increase in parameters and training/inference time due to the hybrid layer}
\label{tab:cost}
\begin{center}
\begin{tabular}{lccccccccc}
\toprule
\multicolumn{1}{c}{} & \multicolumn{1}{c}{\bf OT} & \multicolumn{1}{c}{\bf HI} & \multicolumn{1}{c}{\bf AD} & \multicolumn{1}{c}{\bf HE} & \multicolumn{1}{c}{\bf JA} & \multicolumn{1}{c}{\bf YE} & \multicolumn{1}{c}{\bf DI} & \multicolumn{1}{c}{\bf HO} & \multicolumn{1}{c}{\bf MEAN} \\
\midrule
Parameters & 35.2\% & 34.9\% & 10.4\% & 26.5\% & 33.6\% & 31.9\% & 23.4\% & 27.4\% & 27.9\% \\
Training Time & 35.5\% & 46.2\% & 40.9\% & 43.7\% & 38.0\% & 37.6\% & 45.2\% & 46.5\% & 41.7\% \\
Inference Time & 0.8\% & 3.0\% & 9.0\% & 2.1\% & 0.4\% & 0.9\% & 3.5\% & 3.4\% & 2.9\% \\
\bottomrule
\end{tabular}
\end{center}
\end{table}

The introduction of the proposed hybrid transformer layer has led to significant improvements in accuracy, though it comes with an increase in the number of trainable parameters. Table \ref{tab:cost} details the additional parameters incurred by employing the hybrid architecture over a standard transformer encoder, as well as the corresponding rise in training and inference times for a single batch. Although the increase in parameters is measurable, it is not excessive, with an average rise of $28\%$ and a maximum of just over $35\%$. Furthermore, this increase moderately affects training time, with increments ranging from $35\%$ to $47\%$. As expected, the augmentation does not significantly impact inference time, given that the additional modules operate in parallel. Finally, it is essential to note that, despite the additional parameters, our encoder-only architectures remain significantly more resource-efficient than some previous models, such as TabNet.

\subsection{Suggested hyper-parameters}

\begin{table*}[h]
\caption{Suggested main hyperparameters}
\label{tab:suhyper}
\begin{center}
\begin{tabular}{lllllllll}
\toprule
\multicolumn{1}{c}{\bf Model} & \multicolumn{1}{c}{\bf HI} & \multicolumn{1}{c}{\bf AD} & \multicolumn{1}{c}{\bf OT} & \multicolumn{1}{c}{\bf HE} & \multicolumn{1}{c}{\bf JA} & \multicolumn{1}{c}{\bf YE} & \multicolumn{1}{c}{\bf DI} & \multicolumn{1}{c}{\bf HO} \\
\midrule 
Dropout & 0.25 & 0.1 & 0.25 & 0.25 & 0.25 & 0.25 & 0.1 & 0.125 \\
Embedding Size & 256 & 16 & 192 & 96 & 192 & 128 & 96 & 128 \\
Depth & 4 & 3 & 5 & 7 & 4 & 6 & 4 & 4 \\
\bottomrule
\end{tabular}
\end{center}
\end{table*}

In Table \ref{tab:suhyper}, we list the main hyperparameters of the proposed model for each dataset, corresponding to the experimental results presented. These values might not showcase the absolute best performance, as we opted against exhaustive optimization and did not use black-box optimization techniques that may require hundreds of iterations to provide optimal results. Additionally, in situations with marginally differing results, factors such as model size were also taken into consideration.

\end{document}